\documentclass[runningheads]{llncs}
\usepackage{graphicx}
\usepackage{hyperref}
\usepackage{comment}
\usepackage{todonotes}

\usepackage{paralist}
\usepackage{xspace}
\usepackage{amsmath,amssymb,amsfonts}
\usepackage{subfig}
\usepackage{color}
\usepackage{tikz}
\usepackage{booktabs}
\graphicspath{imgs}
\usepackage{algorithm}
\usepackage{algpseudocode}

\newcommand{\U}{\mathsf{U}\xspace}
\newcommand{\X}{\mathsf{X}\xspace}
\newcommand{\Xw}{\mathsf{X}_w\xspace}
\newcommand{\G}{\mathsf{G}\xspace}
\newcommand{\W}{\mathsf{U}_w\xspace}
\newcommand{\F}{\mathsf{F}\xspace}

\newcommand{\LTL}{LTL\xspace}
\newcommand{\FLTLF}{FLTL$_f$\xspace}
\newcommand{\FLTL}{FLTL\xspace}
\newcommand{\LTLF}{LTL$_f$\xspace}

\DeclareMathOperator{\len}{length}
\algnewcommand{\Downto}{\textbf{ downto }}

\newcommand{\fluent}[1]{\mathtt{#1}\xspace}

\begin{document}
\title{Conformance Checking of Fuzzy Logs against Declarative Temporal Specifications}
\titlerunning{Conformance Checking of Fuzzy Logs}
%
\author{Ivan Donadello\inst{1}\orcidID{0000-0002-0701-5729} \and
Paolo Felli\inst{2}\orcidID{0000-0001-9561-8775} \and
Craig Innes\inst{3}\orcidID{0000-0002-6329-4136} \and 
Fabrizio Maria Maggi\inst{1}\orcidID{0000-0002-9089-6896} \and
Marco Montali\inst{1}\orcidID{0000-0002-8021-3430}}
\authorrunning{I. Donadello et al.}
%
\institute{Free University of Bozen-Bolzano, Bolzano, Italy \\
\email{\ \{ivan.donadello,maggi,montali\}@inf.unibz.it} 
\and
University of Bologna, Bologna, Italy \\
\email{paolo.felli@unibo.it}\\
\and
University of Edinburgh, Edinburgh, UK \\
\email{craig.innes@ed.ac.uk}\\
}
\maketitle              
\begin{abstract}
Traditional conformance checking tasks assume that event data provide a faithful and complete representation of the actual process executions. This assumption has been recently questioned: more and more often events are not traced explicitly, but are instead indirectly obtained as the result of event recognition pipelines, and thus inherently come with uncertainty.
In this work, differently from the typical probabilistic interpretation of uncertainty, we consider the relevant case where uncertainty refers to which activity is actually conducted, under a fuzzy semantics. 
In this novel setting, we consider the problem of checking whether fuzzy event data conform with declarative temporal rules specified as Declare patterns or, more generally, as formulae of linear temporal logic over finite traces (\LTLF). This requires to relax the assumption that at each instant only one activity is executed, and to correspondingly redefine boolean operators of the logic with a fuzzy semantics. Specifically, we provide a threefold contribution. First, we define a fuzzy counterpart of \LTLF tailored to our purpose. Second, we cast conformance checking over fuzzy logs as a verification problem in this logic. Third, we provide a proof-of-concept, efficient implementation based on the PyTorch Python library, suited to check conformance of multiple fuzzy traces at once.  

\keywords{Conformance checking
\and fuzzy event logs
\and fuzzy linear temporal logic over finite traces}
\end{abstract}

\section{Introduction}
\label{sec:intro}

Conformance checking \cite{CDSW18} is one of the cornerstones of process mining. Its purpose is to compare the expected behaviour described by a reference process model, with the observed behaviour recorded inside event logs as a result of actual executions of the process. This is essential to detect deviations and provide insights that can be used to counter sources of non-compliance and/or to improve the process model. In traditional conformance checking, the recorded execution traces are considered to be \emph{crisp}, that is, they provide a faithful and complete representation of what has been effectively executed. 

This strong assumption has been recently questioned: more and more often, process-level events are not explicitly recorded, but are instead implicitly derived from low-level data through event process and recognition pipelines. Two prime settings incarnate this key fact. The first is the one where execution-relevant data are recorded through sensors at a low-granular level. Such sensor-level data consequently need to be processed, correlated, and aggregated, to ultimately infer abstract, process-level events \cite{ZMLK21}. A typical application scenario of this setting is the usage of ambient sensors to obtain temporal information on the activities performed by humans in their everyday life \cite{DiFB23}.

A second, relevant setting is the one where execution-relevant data consist of raw, unstructured multimedia data, such as, most prominently, videos. To extract meaningful, process-level events, machine (deep) learning pipelines are typically employed, training them on videos annotated with time intervals and the events occurring therein, then using these to recognise events/activities on unseen video recordings. This is essential for human-driven processes where domain experts operate in the physical world and do not always interact with information systems. The importance of this setting has been already recognised, leading also to software architectures that facilitate the interconnection of video processing techniques with process mining \cite{KrKR22}. However, we need a new generation of process mining --in particular conformance checking-- techniques reflecting the \emph{epistemic uncertainty} of the traces recognised through these pipelines \cite{Beer23}. 

The typical approach to fill this research gap is by assuming that ``the log is stochastically known'' \cite{Gal23}. This means that uncertainty is interpreted in a \emph{probabilistic} way, by quantifying the likelihood that an uncertain element of the trace can be resolved into a specific value \cite{PeUA21,BoCG22,FGMRW22,FGMRW23}. Under this interpretation, a confidence value $c\in[0,1]$ attached to an activity label $\fluent{a}$ expresses the confidence that $\fluent{a}$ \emph{was indeed the activity being executed} while the event was logged. When multiple activities $\fluent{a}_1,\ldots,\fluent{a}_n$ are attached to the same event, their confidence values $c_1,\ldots,c_n$ sum up to 1, and the interpretation is that the event actually referred to $\fluent{a}_i$ (and only $\fluent{a}_i$) with probability $c_i$.

Interestingly, in the context of event recognition of videos, other interpretations of uncertainty may be needed, depending on the application domain and characteristics of the event recognition pipeline. For example, in recent video datasets for activity recognition \cite{LTRC20,TRSV21}, events are defined splitting the video in segments and, in each segment, \emph{multiple activities may be concurrently executed, each with its own confidence level}. 
Hence, one or more agents may be present in the scene, each being engaged in possibly different activities at once, and the engagement of an agent in an activity may have different degrees of ``intensity''. 

For example, a camera may be used to record the activities conducted in a human-cobot production line, where human operators and cobots cooperatively work on assembling devices. In a segment, the video may record a situation where a cobot holds a piece firmly, while a human operator applies glue on the piece, also helping the cobot to hold it in place. Through a machine learning pipeline, we could extract from this segment an event associated to the three activities $\fluent{cobot\_holds}~(\fluent{CH})$, $\fluent{human\_holds}~(\fluent{HH})$, and $\fluent{human\_glues}~(\fluent{HG})$, respectively coming with values $c_{\fluent{CH}} = 0.9$, $c_{\fluent{HH}} = 0.2$, $c_{\fluent{HG}} = 0.8$.\footnote{Often, these  values collectively sum up to 1. However, this is not necessarily an indication that they conceptually form a probability distribution, but simply  results from the application of a \emph{softmax} function in the machine learning pipeline.} 
These values 
are not indicating that the three activities are mutually exclusive, but rather how much they are being executed, which is akin to a \emph{fuzzy semantics}.
In a \emph{fuzzy log}, the value of an activity captures \emph{how much of that activity} there is in the event that was logged. 
Conformance checking over fuzzy logs containing this type of events consequently call for measuring \emph{to what degree} the traces contained in the log match with the reference process model. 

In this paper, we study this problem for the first time, in the case where the reference model is captured by declarative, temporal rules, specified as patterns in the Declare language \cite{PeSV07,MPVC10} or, more generally, as formulae of linear temporal logic over finite traces (\LTLF)~\cite{DGV13}, which can express all Declare patterns \cite{DeDM14}. 

To obtain a suitable framework for \emph{conformance checking of fuzzy logs against declarative temporal specifications}, we provide a threefold contribution.
The first concerns the definition of a suitable fuzzy counterpart of \LTLF tailored to our purpose. On the one hand, due to (uncertain) true concurrency in the activities associated to events, we have to drop the so-called ``Declare assumption'' that, at every instant, one and only one activity is executed. On the other hand, to deal with fuzzy event data, we need to redefine the state formulae, and in particular the boolean operators, under a fuzzy semantics. We do so by selecting a suitable temporal logic with fuzzy boolean operators \cite{FrigeriPS14}, defining it over \emph{finite traces}. 

As a second contribution, we cast conformance checking over fuzzy logs as a verification problem in this logic, called \FLTLF. Considering the aforementioned example of a production line, we can express in \FLTLF that ``whenever a piece is glued by the human operator while the cobot holds it, then a quality check of the piece must later be conducted'' -- in formulae: $\G ( (\fluent{CH} \land \fluent{HG}) \rightarrow \F (\fluent{QC}))$, where  $\fluent{quality\_check}~(\fluent{QC})$ is an additional activity. When checking conformance, the state formulae in the specification  are fuzzy. E.g., when $\fluent{CH} \land \fluent{HG}$ is evaluated over the event described above, we will obtain a fuzzy truth value of $0.8$.

Interestingly, our approach blurs the distinction between two activities that are executed truly concurrently, for instance by two distinct teams of workers or by the cobot and the human operator, and two activities that, to some degree, are both in execution, 
like the activities of holding and gluing by the operator. For the purpose of conformance checking, this distinction is inconsequential. 
 
The last contribution shows that our framework can be effectively realized. We do so by reporting on an implementation of the approach where fuzzy traces are encoded into tensors, making it possible to employ the PyTorch Python library to effectively check conformance of multiple fuzzy traces at once. Feasibility is assessed through an experimental evaluation on synthetic data, which reveals that conformance degrees for complex \FLTLF formulae can be computed in the order of 
seconds for a log with 100K traces, each containing 3K events.

\section{Related Work}
\label{sec:related}

Traditional conformance checking does not deal with uncertainty: the reference process model induces a (typically infinite) set of model traces, each considered as equally possible, while the observed trace provides a crisp ordering over its constitute events, each linked in a crisp way to a reference activity. A recent, ongoing wave of conformance checking techniques incorporates uncertainty either in the model or in the log. Along the first line, stochastic conformance checking techniques have been proposed, associating probabilities to the model traces, and using such probabilities when checking conformance of an observed trace \cite{BMMP21}, or deriving overall metrics on the conformance of a log as a whole \cite{LABP21,PolK22,LeMM22}. 

Probabilistic conformance checking is tackled also in the declarative spectrum, considering Declare constraints with attached probability conditions, which are combined to detect, for an observed trace, whether it conforms and, if so, whether it  is a likely or outlier variant of the process \cite{AMMP22}. While all these works adopt a probabilistic interpretation of uncertainty, to our knowledge \cite{ZhangGDNLDK22} is the only one tackling a form of fuzzy conformance, where fuzziness is attached to the temporal dimension of a procedural model with explicit flows.

Along the second line, different forms of probabilistic uncertainty have been attached to event logs \cite{Gal23}, considering (crisp) Petri net-based procedural process models, and dealing with uncertainty on the actual timestamp of events and on the referred activities \cite{PeUA21,BoCG22}, as well as these forms combined also with uncertainty on data payloads \cite{FGMRW22}. All these works consider forms of alignments where every (uncertain) log trace corresponds to multiple possible realisations, where each realization is a crisp trace that comes with an overall probability, computed from the probabilities explicitly attached to its constitutive elements.  

Beside the key difference that in our setting we consider declarative temporal specifications as a reference model, the fact that we work on fuzzy events makes the notion of realisation not applicable. In fact, each event refers to all activities, but for each of them, it does so \emph{to some degree}. Finally, we observe that our approach is orthogonal to the one in~\cite{UmiliCG23} that uses fuzzy automata for conformance checking, but is faster as it does not require the expensive construction of an automaton from an \LTLF formula.

\section{Fuzzy Linear Temporal Logic on Finite Traces}
\label{sec:fltlf}


In this section we introduce the logic, called \FLTLF, that we use to formalize specifications to be verified against fuzzy logs, hence casting the conformance checking problem as a verification problem. 
\FLTLF is a variant of the known Fuzzy Linear-time Temporal Logic (\FLTL)~\cite{LK00,FrigeriPS14}, which in turn is an extension of Zadeh Logic~\cite{Zadeh65} with temporal operators, i.e., a fuzzy counterpart of standard \LTL. 
More specifically, the difference with respect to \FLTL is simply that \FLTLF is interpreted over finite traces, here called sequences, and thus can be seen as a fuzzy counterpart of \LTLF. 
Nonetheless, this simple difference requires particular care in handling the end of sequences when defining the semantics of strong and weak temporal operators in the logic and, later, in the implementation. 

Given a finite set of propositional symbols $P$, a  \FLTLF formula $\varphi$ is defined according to the following grammar:
\[
\varphi := \top ~|~ \bot ~|~ p ~|~ \neg \varphi ~|~ \varphi \land \varphi ~|~ \X \varphi ~|~ \varphi \U \varphi
\]
where $p\in P$ and symbols $\top,\bot$ are the \emph{true} and \emph{false} constants, respectively.  
The temporal operators are $\X$ (next) and $\U$ (until).

Intuitively, these formulae are interpreted over finite, non-empty sequences of instants in which a real value in the interval $[0,1]$ is assigned to each propositional symbol in $P$. 
In this paper, we interpret these values as fuzzy values, which are used to specify the degree of truth of each proposition.   
In the next section, we will discuss how these values can be interpreted in the application domain of interest, namely that of conformance checking of fuzzy event logs. 



\begin{definition}
We call \emph{sequence} of length $n$ a vector of functions of the form $\lambda=\langle \lambda_0,\ldots,\lambda_{n-1}\rangle$ where for each  instant $i = 0,\ldots,n{-}1$ the function $\lambda_i : P \mapsto [0,1]$ assigns to each symbol $p\in P$ a real value. For brevity, we denote by $\lambda(p,i)$ the value associated to $p$ at instant $i$, namely $\lambda_i(p)$.   
\end{definition}

For instance, $\lambda(\fluent{a},0)$ is the value associated to  symbol $\fluent{a}$ at instant $0$, namely at the beginning of the sequence, while $\lambda(\fluent{a},1)$ is the value of $\fluent{a}$ in the following instant. 
%
%
%
%
Crucially, we here consider only non-empty sequences of finite length, and denote the last instant of a sequence $\lambda$ of length $n$ by  $last(\lambda)=n{-}1$.

%
%
%

\begin{example}
\label{ex:sequence}
Consider the following sequence $\lambda_{cobot}$, where propositions symbols are inspired to the  human-cobot example from Section~\ref{sec:intro}, namely $P$ is the set $\{ \fluent{CH}, \fluent{HH}, \fluent{HG},\fluent{QC}\}$. In the next section, we define fuzzy log traces and show how to capture them as sequences of this form (see Example~\ref{ex:trace}). Here, there is still no notion of activity label nor log trace, hence these are ordinary proposition symbols with no specific meaning. $\lambda_{cobot}$ is the sequence:
%
%
{
\[
\begin{array}{r l} 
 \langle 
	\{ \fluent{CH} \mapsto 0.6, \fluent{HH} \mapsto 0.5, \fluent{HG} \mapsto 0, \fluent{QC} \mapsto 0  \}  , &
	\{ \fluent{CH} \mapsto 1, \fluent{HH} \mapsto 0.2, \fluent{HG} \mapsto 0.9, \fluent{QC} \mapsto 0  \}  , \\
         \{ \fluent{CH} \mapsto 0.9, \fluent{HH} \mapsto 0.3, \fluent{HG} \mapsto 0.8, \fluent{QC} \mapsto 0  \}  
	& \{ \fluent{CH} \mapsto 0.9, \fluent{HH} \mapsto 0.3, \fluent{HG} \mapsto 0.8, \fluent{QC} \mapsto 0  \}  \\
 \{ \fluent{CH} \mapsto 0.9, \fluent{HH} \mapsto 0.3, \fluent{HG} \mapsto 0, \fluent{QC} \mapsto 0.6 \} 
 & 
  \{ \fluent{CH} \mapsto 0.9, \fluent{HH} \mapsto 0.1, \fluent{HG} \mapsto 0, \fluent{QC} \mapsto 0.9 \} 
\rangle
\end{array}
\]
}

\noindent
%
%
%
In the first instant $i{=}0$, $\fluent{CH}$ has value $0.6$, $\fluent{HH}$ has value $0.5$, $\fluent{HG}$ has value $0$ and  $\fluent{QC}$ has value $0$. 
In the second instant $i{=}1$, $\fluent{CH}$ has value $1$, $\fluent{HH}$ has value $0.2$, $\fluent{HG}$ has value $0.9$ and  $\fluent{QC}$ has value $0$, and so on. 
These values express the degree of truth of the proposition in each instant. 
As the length is $6$, $last(\lambda_{cobot})=5$. 
\end{example}

The evaluation of a \FLTLF formula $\varphi$ over a sequence $\lambda$ at instant $i\geq 0$, denoted by $v(\varphi,\lambda,i)$, is defined as: 

\[
\begin{array}{r c l}
v(\top,\lambda,i) & = & 1 \text{ and } v(\bot,\lambda,i)  =  0; \\ 
v(p,\lambda,i) & = & \lambda(p,i); \\
v(\neg \varphi,\lambda,i) & = & 1 - v(\varphi,\lambda,i);\\
v(\varphi_1 \land \varphi_2,\lambda,i) & = & min \{~ v(\varphi_1,\lambda,i)~,~ v(\varphi_2,\lambda,i) ~\}; \\
v(\X \varphi,\lambda,i) & = & v(\varphi,\lambda,i{+}1) \text{ if } i<last(\lambda) \text{, } 0 \text{ otherwise}; \\
v(\varphi_1 \U \varphi_2,\lambda,i) & = &  max\{~ v(\varphi_2,\lambda,i) ~,~ min \{~ v(\varphi_1,\lambda,i)~,~  v(\varphi_1 \U \varphi_2,\lambda,i{+}1) ~\} \}  \\
& & \text{if } i\leq last(\lambda) \text{, } 0 \text{ otherwise}. \\
\end{array}
\]
where $\varphi_1,\varphi_2$ are \FLTLF formulae and $p\in P$.

%

We say that a \FLTLF formula $\varphi$ \emph{has value $k$} in a sequence $\lambda$ at instant $i\geq 0$ iff $v(\varphi,\lambda,i)=k$.    
Similarly, we say that $\varphi$ has value $k$ in $\lambda$ iff $v(\varphi,\lambda,0)=k$, also written   $v(\varphi,\lambda)=k$. 
Given a \FLTLF formula $\varphi$ and a sequence $\lambda$, the main reasoning task we consider is to determine the value of $\varphi$ in $\lambda$, namely $v(\varphi,\lambda)$. 
%

%
Note that, under this semantics, the excluded middle principle does not hold and formulae such as $\fluent{a} \land \neg\fluent{a}$ become satisfiable with positive values. 

\begin{example}
Considering the sequence $\lambda_{cobot}$ from Example~\ref{ex:sequence}, $v(\fluent{CH} \land \fluent{HH},\lambda_{cobot})=0.5$, which we interpret as the degree to which both $\fluent{CH}$ and $\fluent{HH}$ are true at instant $i=0$. 
The value is therefore obtained as $min\{0.6,0.5\}$. 
Similarly, $v(\X(\fluent{HG}),\lambda_{cobot}) \allowbreak = 0.9$ (which is the value of $\fluent{HG}$ at instant 1), and $v(\X(\neg\fluent{HH}),\lambda_{cobot},1)=0.7$ (which is equal to the complement of the value of $\fluent{HH}$ at instant 2). 
\end{example}

More complex formulae require more involved computations. For example, the value of the formula $\varphi_1 \U \varphi_2$ depends on the degree to which $\varphi_1$ remains true (i.e., has positive value) until $\varphi_2$ is true  at some instant $j\geq i$. If we fix the choice of $j$, it is computed by taking the minimum between the value of $\varphi_2$ at $j$  and (the minimum of) all the values of $\varphi_1$ up to $j{-}1$. The outermost $max$ operator expresses that we take the best choice of $j$, namely the one giving the highest result. This will be the (highest possible) degree of truth of $\varphi_1 \U \varphi_2$. 

\begin{example}
Considering again $\lambda_{cobot}$ from Example~\ref{ex:sequence}, the value of the formula $(\fluent{HH})\U(\fluent{HG})$ is $0.5$, which is obtained by choosing $j{=}1$. 
Although $v(\fluent{HG},\lambda_{cobot},1)=0.9$, it is $v(\fluent{HH},\lambda_{cobot},0)=0.5$ to determine the final value in this case.
Similarly, if we choose $j{=}2$, although $\lambda(\fluent{HG},\lambda_{cobot},2)=0.8$ we would obtain a lower value of  $0.2$, because $min \{~ \lambda(\fluent{HH},\lambda_{cobot},0),\allowbreak \lambda(\fluent{HH},\lambda_{cobot},1)~\}=0.2$. 
In words, the formula is best satisfied when we consider the degree of proposition $\fluent{HG}$ being true at instant $2$ rather than in any other following instant, since up to that point (i.e., in the first instant $i=1$) the value of $\fluent{HH}$ is higher. 
\end{example}

%


Two \FLTLF formulae $\varphi_1,\varphi_2$ are equivalent, written $\varphi_1\equiv\varphi_2$, if and only if they have the same value in every possible sequence. 
%
%
Hence we can easily prove not only that the usual equivalences from propositional logic apply (as in \FLTL), namely   
$\neg\neg \varphi\equiv \varphi$, 
$\neg(\neg\varphi_1\lor \neg\varphi_2)\equiv \varphi_1\land \varphi_2$ and $\neg(\neg\varphi_1\land \neg\varphi_2)\equiv \varphi_1\lor \varphi_2$, 
but also that we obtain the standard abbreviations as in \LTLF in the usual manner, and so define these additional temporal operators: 
\[
\begin{array}{rclrcl}
\F \varphi &\equiv& \top \U \varphi &
\G \varphi &\equiv& \neg \F \neg \varphi \\
\Xw \varphi &\equiv& \X\varphi \lor \neg \X\top \qquad  &
\varphi_1 \W \varphi_2 &\equiv& \varphi_1 \U \varphi_2 \lor \G \varphi_1 
\end{array}
\]
\noindent
namely $\F$ (eventually), $\G$ (always), $\Xw$ (weak next), $\W$ (weak until).  
The meaning of these operators is illustrated below.

For completeness, 
instead of regarding this extended syntax only as syntactic sugar, we give below also an explicit, direct semantics. This shows explicitly how these operators can be seen as a counterpart of those of \LTLF, making their interpretation (discussed below) and implementation (in Section~\ref{sec:implementation}) more clear.  

\[
\begin{array}{r c l}
v(\varphi_1 \lor \varphi_2,\lambda,i) & = & max \{~ v(\varphi_1,\lambda,i) ~,~  v(\varphi_2,\lambda,i) ~\}; \\
v(\Xw \varphi,\lambda,i) & = & v(\varphi,\lambda,i{+}1) \text{ if } i<last(\lambda) \text{, }  1 \text{ otherwise}; \\
v(\F \varphi,\lambda,i) & = &  max\{~ v(\varphi,\lambda,i) ~,~  v(\F \varphi,\lambda,i{+}1) ~\} \text{ if } i\leq last(\lambda) \text{, } 0 \text{ otherwise;} \\
v(\G\varphi,\lambda,i) & = & min \{~ v(\varphi,\lambda,i) ~,~ v(\G\varphi,\lambda,i{+}1) ~\} \text{ if } i\leq last(\lambda) \text{, } 1 \text{ otherwise;} \\
v(\varphi_1 \W \varphi_2,\lambda,i) & = &  max\{ ~v(\varphi_2,\lambda,i) ~,~ min \{~ v(\varphi_1,\lambda,i)~,~  v(\varphi_1 \W \varphi_2,\lambda,i{+}1) ~\}\}  \\
& & \text{if } i\leq last(\lambda) \text{, } 1 \text{ otherwise}. \\
\end{array}
\]

The proof that the equivalences hold under this explicit semantics is direct and is omitted for brevity. 

Note that the semantics of these additional temporal operators is given in a recursive way for uniformity with the others, although for $\F$ and $\G$ this can be stated more directly. 
Intuitively, the value of a formula $\F\varphi$ in a sequence, at some instant, is the highest degree to which $\varphi$ is true either at that instant or at some future instant. It can therefore be computed as $v(\F\varphi,\lambda,i)=max\{ v(\varphi,\lambda,i), \ldots, v(\varphi,\lambda,last(\lambda)) \}$.
Conversely, the value of a formula $\G\varphi$ at some instant 
is the lowest degree to which $\varphi$ is true from that instant until the the end of the sequence, namely 
$v(\G\varphi,\lambda,i)=min\{ v(\varphi,\lambda,i), \ldots, v(\varphi,\lambda,last(\lambda)) \}$.

%
%
%

\begin{example}
\label{ex:running_formula_computed}
Consider the formula $\G ( (\fluent{CH} \land \fluent{HG}) \rightarrow \F (\fluent{QC}))$ (discussed in Section~\ref{sec:intro} for the cobot-human example). 
The value obtained evaluating this formula on the sequence $\lambda_{cobot}$ from Example~\ref{ex:sequence}, computed as the minimum value of formula $\varphi=(\fluent{CH} \land \fluent{HG}) \rightarrow \F (\fluent{QC})$ along the sequence, is $0.9$. 
Indeed, $v(\F (\fluent{QC}),\lambda_{cobot},i)=0.9$ for each instant $i\in[0,6]$, because this is the value of $\fluent{QC}$ at instant $6$ and also the highest. 
At instants $i=0,4,5$, the value of $\fluent{HG}$ is $0$ and thus the antecedent $(\fluent{CH} \land \fluent{HG})$ in $\varphi$ has value $0$. 
Therefore the value of $\varphi$ is maximal (i.e., $1$) at these instants (since $\varphi_1\rightarrow \varphi_2\equiv \neg\varphi_1 \lor \varphi_2$). 
In instants $i=1,2,3$, the value of the antecedent $(\fluent{CH} \land \fluent{HG})$ is $0.9$, $0.8$, $0.8$, respectively, so its negation has value at most $0.2$, which is less than $0.9$. Hence $\varphi$ has value $0.9$ in these instants.  
As we are taking the minimum value along the trace, the value of the formula is $0.9$. 
\end{example}

The operators $\Xw$ and $\W$ are the fuzzy counterpart of these in \LTLF, and represent a weakening of the next and until.  
%
%
In \LTLF, the weak next operator does not require the next instant to exist, and $\Xw\varphi$ is always true at the last state of a trace. Accordingly, here $v(\Xw \varphi, \lambda,last(\lambda))=1$ (maximal degree of truth) no matter what the values in $\lambda$ are. 
For $0\leq i<last(\lambda)$, $v(\Xw \varphi,\lambda,i)=v(\X \varphi,\lambda,i)$. 

Regarding the weak until, this operator in \LTLF does not require the right-hand side formula to ever become true, as long as the left-hand side is always true.  
Here, as $\varphi_1 \W \varphi_2 \equiv \varphi_1 \U \varphi_2 \lor \G \varphi_1$, we have $v(\varphi_1 \W \varphi_2,\lambda,i)= max \{ v(\varphi_1 \U \varphi_2,\lambda,i) , \allowbreak v(\G\varphi_1,\lambda,i) \}$. 
%
Intuitively, if the degree to which $\varphi_2$ will eventually be true is lower than the degree of $\varphi_1$ being true at each instant along the entire sequence (i.e., its minimum value), then the value of $\G\varphi_1$ is selected instead of that of $\varphi_1 \U \varphi_2$. 
%
%

We note that our logic collapses to standard \LTLF in the special case in which only crisp values (namely $0$ or $1$) are present in sequences. We call sequences in this subclass crisp sequences.  
We recall that \LTLF formulae are interpreted on finite traces, namely sequences of states of the form $\tau=s_0,s_1,\ldots, s_{n-1}$ where each $s_i\subseteq 2^P$ is a set of propositional symbols that are true in the $i$-th state. One writes $\tau\models \varphi$ to denote that the \LTLF formula $\varphi$ is satisfied by $\tau$~\cite{DGV13}. 
Given a crisp sequence $\lambda$, it is trivial to transform it into a standard \LTLF trace, which we denote by $trace(\lambda)$: $p\in s_i$ iff $\lambda(p,i)=1$, for each $i\in [0,last(\lambda)]$. 
Further, let us denote by LTL$_f$($\varphi$) the standard \LTLF formula equal to $\varphi$ (as these two languages share the same syntax).  
Then, the following result holds:

\begin{theorem}
\label{th:byProduct}
Given a crisp sequence $\lambda$ and a \FLTLF formula $\varphi$, then $v(\varphi, \lambda)\in\{0,1\}$ and $v(\varphi, \lambda)=1$ iff $trace(\lambda)\models\text{LTL}_f(\varphi)$.  \end{theorem}

    The proof is trivial. 
    As in Boolean logic, the standard semantics of Boolean connectives can be given by using complement (for negation), $min$ (for conjunction) and $max$ (for disjunction), and $1/0$ for $\emph{true}/\emph{false}$. 
    For temporal operators, the semantics of each operator in one logic  can be immediately reduced to the other, and the same equivalences hold. 
    %
%
The analogous result is already known to hold for \FLTL and the variants of fuzzy-time logics introduced in \cite{FrigeriPS14}. 

As commented in the conclusions, the result above is not only important in order to show that our semantics behaves as expected in the crisp setting, but also that our implementation approach can be readily employed to evaluate standard \LTLF formulae over crisp finite traces. 

However, note that, as for \FLTL, \FLTLF is not adequate to represent the vagueness in the temporal dimension: while Boolean connectives are fuzzy, the temporal operators retain a crisp semantics, so we cannot express statements that capture vagueness in the temporal dimension. 
For this reason, in future work, we plan to look at the application of (a finite-trace version of) more involved logics such as the variants of Fuzzy-time \LTL~\cite{FrigeriPS14}, to express fuzzy temporal specifications involving operators such as `soon' and  `almost always'.

\section{Conformance Checking of Fuzzy Logs}
\label{sec:conformance}

We now introduce the application domain and main reasoning task, namely the conformance checking task for fuzzy event logs of business processes. 
Specifically, we consider a form of uncertain logs akin to those known from the literature of conformance checking with uncertainty~\cite{BoCG22,FGMRW22,FGMRW23}. 

Importantly, however, rather than interpreting the values associated to activity labels in the event log as stochastic information, as done in~\cite{BoCG22}, or as confidence in the correctness of the logged data, as done in~\cite{FGMRW22}, we assume that these were logged as fuzzy values. 
In other words, the values associated to an activity label in each log event expresses to what degree the activity was in execution when the event was recorded. 
For the same event in a trace, more than one activity can have a positive value associated, so that the usual ``Declare assumption" \cite{Mont10,FionG18} does not hold in this framework. 
%


%
For simplicity, as our approach and analysis task is not concerned with timestamps nor case and event IDs, we keep the formalization of fuzzy log traces as lean as possible, focusing primarily on the activity labels associated to each event, and on their fuzzy values.  
Of course, richer representations are also possible. 

\begin{definition}
Given a set of activity labels $A$, a fuzzy log trace is a finite sequence of events $\sigma=E_0, E_1,\ldots E_n$ where each event $E_i$ is a set of couples $\langle a, c \rangle$ where $a\in A$ and $c\in [0,1]$ is a real value associated to $a$, for $i\in[1,n]$. 
\end{definition}

A fuzzy log is a finite set of fuzzy log traces, of varying length.

Intuitively, a couple $\langle a, c \rangle$ in some event $E_i$ encodes the fact that for instant $i$ the activity $a\in A$ was recorded by the logging system with fuzzy value $c$. 
We do not require that $\sum_{\langle a, c \rangle\in E_i} c=1$ for each $i\in[1,n]$, although this is allowed.

It is obvious that we can represent a fuzzy log trace of the form $\sigma=E_0, E_1,\ldots E_n$ as a sequence $\lambda$ as in the previous section of length $n{+}1$: 
\begin{compactitem}
 	\item $P=A$, namely we take activity labels as propositional symbols;
	\item $\lambda(a,i)=c$ for each couple $\langle a, c \rangle$ in event $E_i$, for $i\in[0,n]$;
	\item $\lambda(a,i)=0$ for each activity $a$ not appearing in event $E_i$, for $i\in[0,n]$;
\end{compactitem}

\noindent
We denote this by writing $\lambda=seq(\sigma)$. 

\begin{example}
\label{ex:trace}
Consider the sequence $\lambda_{cobot}$ from Example~\ref{ex:sequence}. 
We can immediately see that it corresponds to the following fuzzy log trace $\sigma_1$, i.e., $\lambda_{cobot}=seq(\sigma_1)$, in a fuzzy log for the cobot-human application introduced in Section~\ref{sec:intro}:
\[
{
\begin{array}{r l} 
\sigma_1=\langle~ 
	\{ \langle \fluent{CH},  0.6 \rangle, \langle \fluent{HH},  0.5 \rangle \}  , &
	\{ \langle \fluent{CH},  1 \rangle, \langle \fluent{HH} , 0.2 \rangle, \langle \fluent{HG}, 0.9 \rangle \} ,  \\
 	\{ \langle \fluent{CH},  0.9 \rangle, \langle \fluent{HH},  0.3 \rangle, \langle \fluent{HG},  0.8 \rangle \}  , &
	\{ \langle \fluent{CH},  0.9 \rangle, \langle \fluent{HH},  0.3 \rangle, \langle \fluent{HG},  0.8 \rangle \}  ,  \\
 \{ \langle \fluent{CH},  0.9 \rangle, \langle \fluent{HH} , 0.3 \rangle, \langle \fluent{QC} , 0.6 \rangle \} , &
 \{ \langle \fluent{CH},  0.9 \rangle, \langle \fluent{HH} , 0.1 \rangle, \langle \fluent{QC} , 0.9 \rangle \}  
~\rangle
\end{array}
}
\]

Unlike in Example~\ref{ex:sequence}, we can finally interpret this fuzzy log trace within the running example as follows (recall that $\fluent{CH}$, $\fluent{HH}$, $\fluent{HG}$ and $\fluent{QC}$ are abbreviations for activity labels $\fluent{cobot\_holds}$, $\fluent{human\_holds}$, $\fluent{human\_glues}$ and $\fluent{quality\_check}$). 

In the first event $E_0$, activity $\fluent{CH}$ has value $0.6$, $\fluent{HH}$ has value $0.5$, $\fluent{HG}$ has value $0$ and  $\fluent{QC}$ has value $0$:  both the cobot and the human operator are holding the piece (with different degrees), but neither gluing nor checking is logged. 
In the second event $E_1$, the cobot is clearly holding the piece (maximal value), the human is somewhat holding the piece while applying glue, and still there is no check performed.  
The following events $E_2,E_3,E_4$ have similar interpretation. 
In the last two events $E_5$ and $E_6$ the activity $\fluent{quality\_check}$ is associated to high, increasing values, which means that the human performed this activity to an increasing degree. In these two instants, and especially in $E_6$, the human is almost not holding the piece anymore. Note how we have used here terms such as `almost', `high', `somewhat', which are fuzzy concepts. 

The value of the formula discussed in Section~\ref{sec:intro}, namely $\G ( (\fluent{CH} \land \fluent{HG}) \rightarrow \F (\fluent{QC}))$, evaluated on the sequence $\lambda_{cobot}=seq(\sigma_1)$, is discussed in Example~\ref{ex:running_formula_computed} and is equal to $0.9$. 
This indicates that the specification is almost completely true in the fuzzy log trace. 
As explained also in Example~\ref{ex:running_formula_computed}, in some instant (corresponding to log events) the conjunction $\fluent{CH} \land \fluent{HG}$ has a high value, hence the cobot and the human are indeed considerably cooperating in holding the work piece, which is a strong indication that some quality check must be performed. Such a check is performed with high value at instant $6$, as witnessed by the high degree  of truth of the eventuality $\F (\fluent{QC})$. An even higher value for the implication, equal to $1$, is in fact obtained at some instants (when the human is not gluing at all, making the specification trivially true), but the $\G$ operator demands that the minimum value along the sequence is returned. 
\end{example}


\begin{definition}
	Given a fuzzy log trace $\sigma$ with activity labels $A$ and a \FLTLF formula $\varphi$ with propositional symbols from the same set $A$, the conformance checking task is to evaluate $\varphi$ over $seq(\sigma)$, namely compute $v(\varphi, seq(\sigma))$. 
\end{definition}

Differently from other frameworks for conformance checking for business processes, rather than considering an explicit process model (given, e.g., as a Petri net), we are computing to what degree a fuzzy log trace conforms to a declarative process specification captured by a \FLTLF formula, namely, the degree of truth for that formula in the log trace. 
This is in line with the literature in BPM in which declarative  reference models are given by resorting to a set of declarative, temporal rules, specified as patterns in the Declare language~\cite{PeSV07,MPVC10} or, more generally, as formulae in \LTLF, which can express all Declare patterns \cite{DeDM14}.

As already pointed out, there is a distinctive trait when adopting Declare in our setting: the ``Declare assumption'', dictating that at every instant, exactly (or at most) one activity can be executed, does not hold anymore. This is in principle not an issue, as Declare patterns are well-defined even when multiple activities can be executed in the same instant \cite{Mont10}. In fact, the semantics of Declare constraints remains unaltered. However, what changes is how constraints interact with each other, as shown in the next example.

\begin{example}
\label{ex:declare-assumption}
Consider a Declare specification that includes three constraints $\varphi_1 = \textsf{chain-}\textsf{response}(\fluent{a},\fluent{b})$, $\varphi_2 = \textsf{chain-response}(\fluent{a},\fluent{c})$, and $\varphi_3 = \textsf{neg-chain-response}(\fluent{a},\fluent{d})$. Recall that pattern $\textsf{chain-response}(x,y)$ indicates that whenever $x$ is executed, then $y$ must be executed in the next instance, and pattern $\textsf{neg-chain-response}(x,y)$ dictates that whenever $x$ is executed, then the next executed activity cannot be $y$. In \LTLF, we thus get $\varphi_1 = \G (\fluent{a} \rightarrow \X \fluent{b})$,  $\varphi_2 = \G (\fluent{a} \rightarrow \X \fluent{c})$, and  $\varphi_3 = \G (\fluent{a} \rightarrow \neg \X \fluent{d})$. 
Under the Declare assumption, only traces that do not contain $\fluent{a}$ would be accepted, as $\varphi_1$ and $\varphi_2$ contradict each other in the presence of $\fluent{a}$ (the first requiring $\fluent{b}$ next, the second requiring $\fluent{c}$ instead). Also, $\varphi_3$ would be redundant with $\varphi_1$ (resp., $\varphi_2$), since requiring a $\fluent{b}$ (resp., $\fluent{c}$) next automatically excludes the possibility of doing any other activity, including $\fluent{d}$.
Without the Declare assumption, the specification indicates that whenever $\fluent{a}$ is currently executed, in the next instant we require the concurrent execution of multiple activities including $\fluent{b}$ and $\fluent{c}$, and excluding $\fluent{d}$. 
\end{example}

Over fuzzy logs, a Declare specification like the one in Example~\ref{ex:declare-assumption} should be interpreted under the fuzzy semantics of \FLTLF. In addition, recall that the Declare assumption can itself be specified in \LTLF \cite{Mont10}: given a set $A$ of activities, it is the conjunctive formula $\G (\bigwedge_{a_i,a_j \in A,i<j} \neg (a_i \land a_j))$. Therefore, also in our framework, the modeller can decide to add such formula explicitly  (also in this case, retaining a fuzzy interpretation for its semantics).

As customary in declarative conformance checking \cite{CDSW18} and monitoring \cite{DDMM22}, when assessing conformance with a  specification, one may be interested in obtaining insights on the entire specification, as well as on its constitutive constraints. This can be directly tackled in our approach. In fact, as Declare specification are conjunctions of constraints, we can obtain a constraint-level feedback by computing the fuzzy value of each constraint, and the specification-level one is directly computed by taking the minimum of those values.

\section{Implementation}
\label{sec:implementation}
An important contribution of the paper is an efficient satisfiability checker for the \FLTLF semantics described in Section~\ref{sec:fltlf}. We implemented this checker by using the PyTorch Python library\footnote{\url{https://pytorch.org/}} for two main reasons: (i) its basic operations are multi-threaded, thus allowing a fast parallel execution of the code; (ii) PyTorch is a library for implementing Deep Learning algorithms, therefore, a satisfiability checker for \FLTLF written in PyTorch can be easily integrated in the loss function of Neural Networks for a Neuro-Symbolic system for, e.g., next event prediction from videos. The parallelism supported by the multithreading requires a different characterization of the log from the traditional log format as a ``flat'' set of traces that are sequences of ordered events. 

To this extent, we encode a (fuzzy) log as a tensor (or $n$-dimensional array) with three dimensions, called \emph{(fuzzy) log tensor}. One dimension indexes the activity labels, one dimension indexes the ordered events (in correspondence to time instants), and the last dimension indexes the traces of the log. Each cell contains the fuzzy value for a given activity in an event belonging to a trace. 

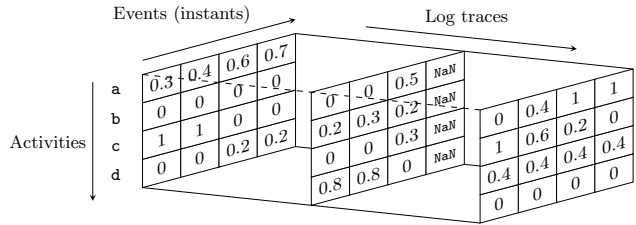
\begin{figure}[t]
\begin{center}
\resizebox{0.7\textwidth}{!}
{
\begin{tikzpicture}[x=(15:.7cm), y=(90:.5cm), z=(354:.5cm), >=stealth]

\def\aaa{{
{{0,0,0.2,0.2},{1,1,0,0},{0,0,0,0},{0.3,0.4,0.6,0.7}},
{},{},{},{},{},
{{0.8,0.8,0, "\scriptsize{$\fluent{NaN}$}"},{0,0,0.3,"\scriptsize{$\fluent{NaN}$}"},{0.2,0.3,0.2,"\scriptsize{$\fluent{NaN}$}"},{0,0,0.5,"\scriptsize{$\fluent{NaN}$}"}},
{},{},{},{},{},
{{0,0,0,0},{0.4,0.4,0.4,0.4},{1,0.6,0.2,0},{0,0.4,1,1}}
}};

\draw (0, 0, 0) -- (0, 0, 12) (4, 0, 0) -- (4, 0, 12);
\foreach \z in {0, 6, 12} \foreach \x in {0,...,3}
  \foreach \y in {0,...,3} 
    %
    %
    \pgfmathsetmacro{\val}{\aaa[\z][\y][\x]}
    \filldraw [fill=white] (\x, \y, \z) -- (\x+1, \y, \z) -- (\x+1, \y+1, \z) --
      (\x, \y+1, \z) -- cycle (\x+.5, \y+.5, \z) node [yslant=tan(15)] 
      {\val};

\draw [dashed] (0, 4, 0) -- (0, 4, 12);
\draw [] (4, 4, 0) -- (4, 4, 12);
\draw [->] (0, 4.5, 0)  -- (4, 4.5, 0)   node [near end, above left] {Events (instants)};
\draw [->] (-1.3, 4.2, 0)  -- (-1.3, 0, 0)   node [midway, left] {Activities};
\draw [<-] (4, 4.5, 10) -- (4, 4.5, 2.5) node [near end, above right] {Log traces};
\node (output) at (-0.7,3.7) {$\fluent{a}$};
\node (output) at (-0.7,2.7) {$\fluent{b}$};
\node (output) at (-0.7,1.7) {$\fluent{c}$};
\node (output) at (-0.7,0.7) {$\fluent{d}$};
\end{tikzpicture}
}
\end{center}
\caption{A fuzzy log represented as a tensor where each log trace is a vertical slice, holding the fuzzy values for each activity label, for each event.}
\label{fig:LogCube}
\end{figure}

Figure~\ref{fig:LogCube} shows an example of fuzzy log tensor with 4 activity labels ($\fluent{a}$, $\fluent{b}$, $\fluent{c}$ and $\fluent{d}$), 3 traces with up to 4 events each.
In this representation, each log trace is therefore a vertical slice, holding the fuzzy values for each activity label, for each event. 
E.g., activity $\fluent{a}$ is given an increasing value along the left-most trace in the figure, whereas $\fluent{b}$ constantly has value $0$, and so on. Log traces normally have varying length, so a padding is needed with, for example, $\mathsf{NaN}$ values.
In the figure, the trace in the middle has length 3 and is padded in the last position. 

Computing the satisfiability value for a given \FLTLF formula given a fuzzy log tensor can be performed by applying the \FLTLF semantics as explained in Section~\ref{sec:fltlf}. Therefore, the evaluation of a constraint against a log trace (a slice from the fuzzy log tensor) can be represented with a tree in which the leaves correspond to the evaluation of the propositional symbols, the intermediate nodes correspond to the evaluation of either Boolean or temporal operators, and the root corresponds to the evaluation of the whole formula. 
%
This corresponds to a computational graph where the internal nodes are basic PyTorch operations and the input in the leaves are horizontal slices of the fuzzy tensor log that contain the fuzzy values of a given activity for all the traces and events. In this way, the traces (or, rather, their corresponding sequences) can be processed in parallel. 

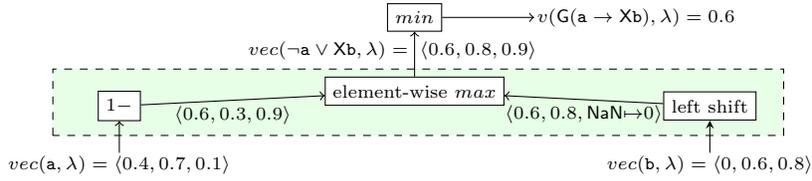
\begin{figure}[t]
\begin{center}
\resizebox{0.9\textwidth}{!}
{
\begin{tikzpicture}[inner sep=3pt,-stealth]\scriptsize
\draw[dashed,fill=green!10] (1.1,0.4) rectangle (11,1.3);

\node[inner sep=1pt] (pc) at (2,-0) {$vec(\fluent{a}, \lambda) = \left<0.4, 0.7, 0.1\right>$};
\node[inner sep=1pt] (lf) at (10,-0) {$vec(\fluent{b}, \lambda) = \left<0, 0.6, 0.8\right>$};
\node[rectangle,fill=white,draw] (neg) at (2,0.8) {$1-$};
\draw[->] (pc) -- (neg);

\node[rectangle,fill=white,draw] (x) at (10,0.8) {left shift};
\draw (lf) -- (x);

\node[rectangle,fill=white,draw] (or) at (6,1) {element-wise $max$};
\node[rectangle,fill=white,draw] (g) at (6,2) {$min$};
\draw[->] (neg) -- (or) node [midway,yshift=-.2cm] {$\langle 0.6, 0.3, 0.9 \rangle$};
\draw[->] (x) -- (or) node [midway,yshift=-.2cm] {$\langle 0.6, 0.8, \mathsf{NaN}{\mapsto} 0 \rangle$} ;
\draw[->] (or) -- (g) node [midway,xshift=-.3cm,yshift=.05cm] {$vec(\neg \fluent{a} \vee \X\fluent{b},\lambda) =\; \langle 0.6, 0.8, 0.9 \rangle$};
\node[inner sep=1pt] (output) at (9,2) {$v(\G(\fluent{a} \rightarrow \X\fluent{b}), \lambda) = 0.6$};
\draw[->] (g) -- (output);
\end{tikzpicture}
}
\end{center}
\caption{Example of computation of $v(\G(\fluent{a} \rightarrow \X\fluent{b}),\lambda)$ where $\lambda=seq(\sigma)$ is the sequence for the log trace $\sigma=\langle \{ \langle\fluent{a}, 0.4\rangle \}, \{ \langle\fluent{a}, 0.7\rangle, \langle\fluent{b}, 0.6 \rangle \}, \{ \langle \fluent{a}, 0.1 \rangle, \langle\fluent{b}, 0.8 \rangle \} \rangle $.}
\label{fig:SateExample}
\end{figure}

Figure~\ref{fig:SateExample} shows such a tree/computational graph for the specification $\G(\neg \fluent{a} \vee \X\fluent{b})$ which is equivalent to $\G(\fluent{a} \rightarrow \X\fluent{b})$. For clarity, we just show a single sequence as input even though, in the implementation, traces are processed in parallel. 

In the figure, we denote by $vec(\varphi,\lambda)$ the vector of fuzzy values obtained by evaluating $\varphi$ on each instant of $\lambda$, namely $\langle v(\varphi,\lambda,0),\ldots,v(\varphi,\lambda,last(\lambda)) \rangle$. First, these vectors are computed for $\fluent{a}$ and $\fluent{b}$, then the values of the former are complemented (hence obtaining the vector of values of $\neg\fluent{a}$) while, in parallel, the values of the latter are shifted of one position to the left, hence obtaining the values of $\X\fluent{b}$. Here, according to the semantics of the strong next, the last value is set to $0$ (indeed, the way we replace padding $\fluent{NaN}$ values depends on the temporal operator being evaluated). Next, an element-wise max gives the vector of values for the disjunction $\neg \fluent{a} \vee \X\fluent{b}$. Finally, the last step implements the semantics of the $\G$ operator, returning the minimum value in the vector.

The \FLTLF semantics for the until operator is defined in Section~\ref{sec:fltlf} in a recursive way, therefore an ad-hoc efficient algorithm has to be implemented. To this end, we implemented a dynamic programming algorithm (see Algorithm~\ref{alg:until}), following the work in~\cite[Section 3.1]{BartocciDDFMNS18} for Signal Temporal Logics, that runs in linear time in the number of events/instants in the fuzzy log tensor. 

\begin{algorithm}[t]
	\caption{The \textsc{Until} evaluation algorithm}
	\label{alg:until}
	\begin{algorithmic}[1]
        \Require $\lambda$, $\varphi_1$, $\varphi_2$
		\State $vec_{\varphi_1} \gets vec(\varphi_1, \lambda)$
        \State $vec_{\varphi_2} \gets vec(\varphi_2, \lambda)$
        \State $vec_\fluent{U} \gets \langle 0, \ldots, 0 \rangle$ \Comment{$vec_\fluent{U}$ has the same length of $vec_{\varphi_1}$ and $vec_{\varphi_2}$}
        \State $vec_\fluent{U}[-1] \gets \min(vec_{\varphi_1}[-1], vec_{\varphi_2}[-1]))$ \Comment{-1 indicates the last array position}
        \For{$i \gets \len(vec_\fluent{U}) -1 \Downto 1$} \Comment{Array indices start from 1}
            \State $vec_\fluent{U}[i] \gets \max(vec_{\varphi_2}[i], \min(vec_{\varphi_1}[i], vec_\fluent{U}[i + 1]))$ \label{line:Min}
        \EndFor
        \State \Return $vec_\fluent{U}[1]$
	\end{algorithmic}
\end{algorithm} 

Due to padding, vectors $vec_{\varphi_1}$ and $vec_{\varphi_2}$ derived from the input can contain $\mathsf{NaN}$ values. Thus, the $\min$ and $\max$ operations (used in line \ref{line:Min}) return a $\mathsf{NaN}$ value if two $\mathsf{NaN}$ values are compared otherwise the non-$\mathsf{NaN}$ value if a non-$\mathsf{NaN}$ value and a $\mathsf{NaN}$ value are compared. We stress that the algorithm shows the computation for just a single log trace, but the implementation considers slices of activity labels (appearing in $\varphi_1$ or $\varphi_2$) to process multiple traces in parallel.

This implementation allows a fast computation of the satisfiability checking of a fuzzy log, as commented in the next section. However, as a byproduct (see Theorem~\ref{th:byProduct}), we obtained also a fast satisfiability checker for standard \LTLF specifications over crisp logs. The source code is available online at~\url{https://github.com/ivanDonadello/Fuzzy-LTLf-Conformance-Checking}.

\section{Evaluation}
\label{sec:evaluation}

To assess the performance of the implemented conformance checker we sampled the checking times of randomly synthetic fuzzy logs for a set of \FLTLF formulae. Note that the conformance task amounts to a form of query answering, where the query is the \FLTLF formula to be checked, the database is a set of traces, and the answer is a corresponding set of truth values. In this respect, it is of particular relevance \cite{Vardi82} to focus on the data complexity, that is, on the performance when the size of the log increases. In fact, the size of the log is much larger than that of the formula to be evaluated. In addition, in the case of Declare, where multiple formulae are conjoined in a possibly large specifications, one can concurrently evaluate each single (small) formula separately from the others, and then simply take the minimum value to compute that of the entire specification.

Synthetic logs are directly generated in tensorial format by creating a tensor of random fuzzy values, with a varying number of traces and events per trace. These dimensions are varied to test the efficiency of the checker over large event logs, where the size of a log is given by the total number of events it contains. 
%
%
We increase the number of traces from 100 to $10^5$ at a base-10 logarithmic step, and the length of each trace from 500 to 3000 events, with deltas of 500 units at each step. In this way, the size of the largest log is $3000\cdot10^5 = 3\cdot10^8$ total events. 
Notice that the standard benchmark logs used in Process Mining tasks have at most $\approx 10^6$ total events, see~\cite[Table 6]{TeinemaaDRM19}. 
Regarding the \FLTLF formulae, we tested the checker over formulae employing a single temporal operator, namely $\X\fluent{a}, \F\fluent{a}, \G\fluent{a}, \fluent{a}\U\fluent{b}$, and a more complex formula $(\fluent{a}\U\fluent{b}) \wedge \G(\fluent{a} \rightarrow \X\fluent{b}) \wedge \G(\fluent{b} \rightarrow \X\fluent{c})$. The set $P$ of propositional symbols is equal to the set of propositional symbols contained in the \FLTLF formulae, that is $P = \{\fluent{a}, \fluent{b}, \fluent{c}\}$. The experiments were performed on a MacBook Pro M1, 16GB RAM, 8 cores.
\begin{figure}[t]
\begin{tabular}{cc}
  \includegraphics[width=0.5\textwidth]{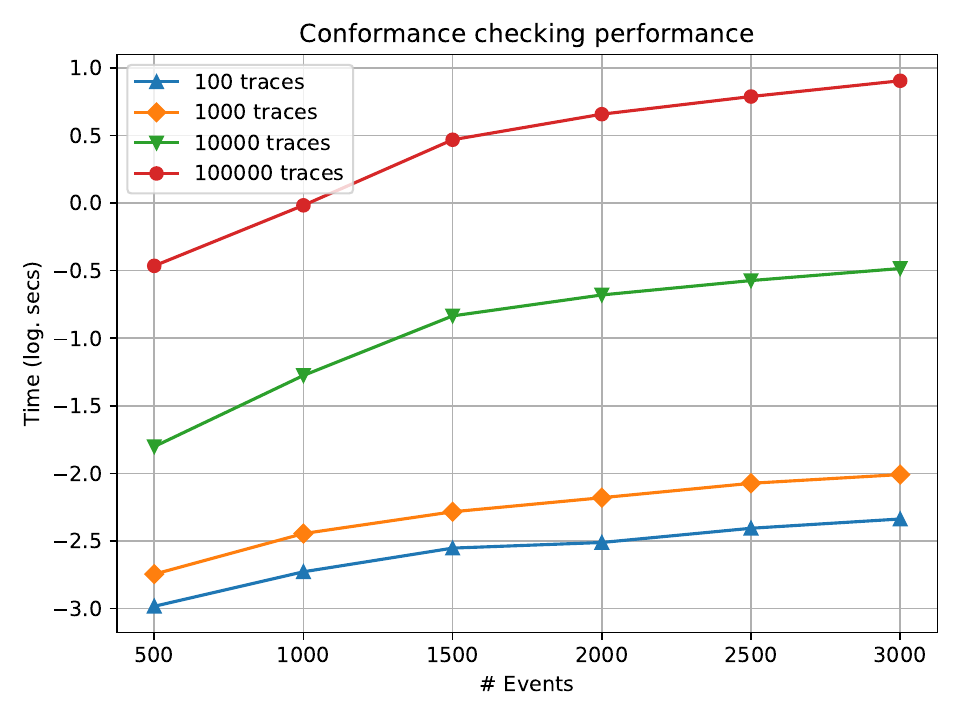} &   \includegraphics[width=0.5\textwidth]{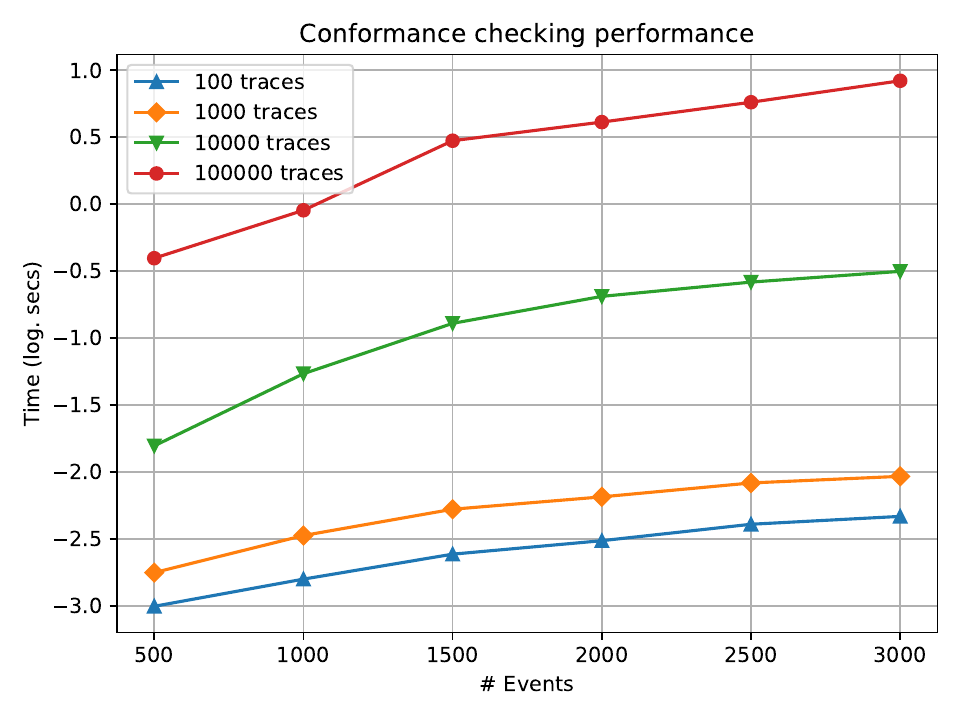} \\
(a) $\F\fluent{a}$ & (b) $\G\fluent{a}$ \\[6pt]
 \includegraphics[width=0.5\textwidth]{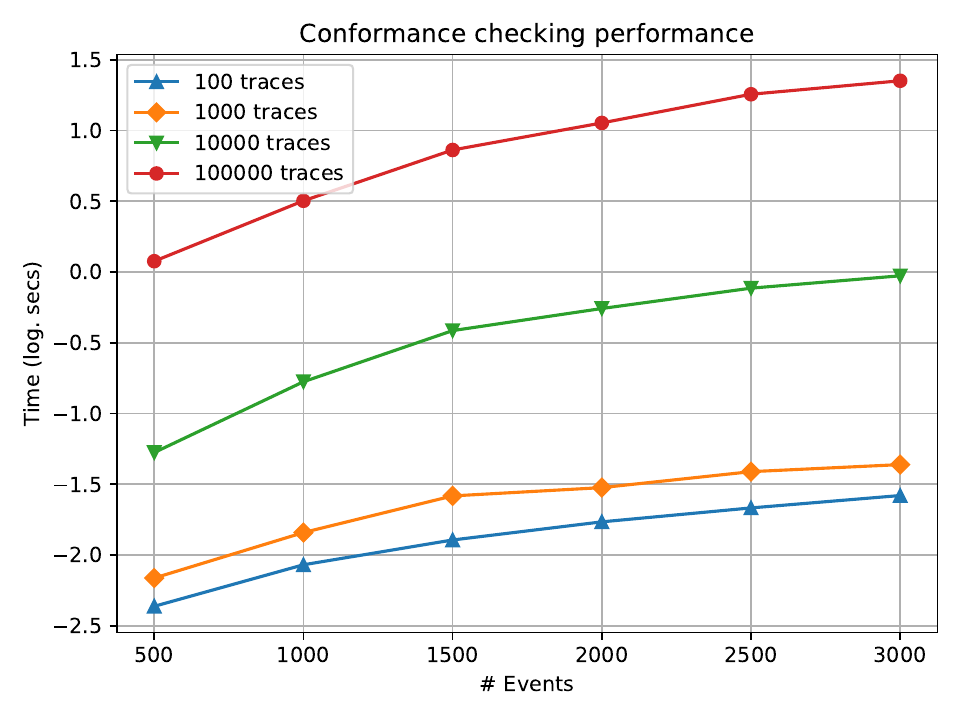} &   \includegraphics[width=0.5\textwidth]{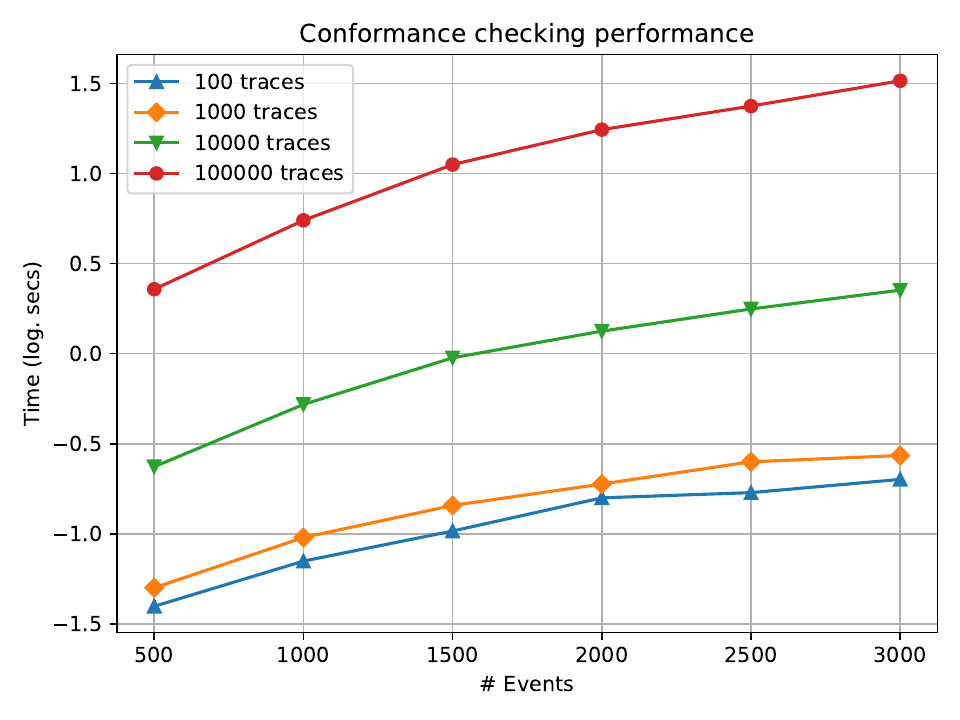} \\
(c) $\fluent{a}\U \fluent{b}$ & (d) $(\fluent{a}\U\fluent{b}) \wedge \G(\fluent{a} \rightarrow \X\fluent{b}) \wedge \G(\fluent{b} \rightarrow \X\fluent{c})$ \\[6pt]
\end{tabular}
\caption{Computational logarithmic times of the conformance checking.}
\label{fig:res}
\end{figure}

Figure~\ref{fig:res} shows the execution times in logarithmic scale as it results in a better visualization than the linear scale. The results for the $\X\fluent{a}$ formula are not plotted as the checker just reports the fuzzy value almost instantaneously. 
We observe a curve with a logarithmic-like shape suggesting a polynomial trend of the execution times when the number of events in each trace increases.

\begin{table}[t]
\centering
\small
\begin{tabular}{@{}lccccc@{}}
\toprule
\#Traces       & $\X\fluent{a}$ & $\F\fluent{a}$ & $\G\fluent{a}$ & $\fluent{a}\U\fluent{b}$ & Complex formula \\ \midrule
100    & 0.00003 & 0.00462 & 0.00492 & 0.02726  & 0.20269  \\
1000   & 0.00049 & 0.00980 & 0.01020 & 0.04713  & 0.27767  \\
10000  & 0.00332 & 0.33113 & 0.34517 & 0.94037  & 2.21606  \\
100000 & 0.02676 & 8.28548 & 8.25296 & 21.34241 & 32.23773 \\
\bottomrule
~\\
\end{tabular}
\caption{Execution times (in secs) of the conformance checking task.}
\label{tab:resBiggestLog}
\end{table}

We also report the linear execution times (in Table~\ref{tab:resBiggestLog}) for each logs by fixing the number of events to the maximum value of $3000$ for each \FLTLF formula. A linear trend in the execution times can be noticed. Moreover, the single numeric values indicates a very efficient performance as the modal operators $\X$, $\F$ and $\G$ are checked in a fast way: they require to just check the activities in the next event (for $\X$), or execute a single maximum/minimum PyTorch operation over the rows of a slice (i.e., a trace) containing the fuzzy values of a given activity along that trace, for each trace in the log (for $\F$ and $\G$, respectively). 
%
%
The most time consuming modal operator is $\U$, which requires more than a single PyTorch operation over an array -- see Algorithm~\ref{alg:until}. 
Nonetheless, the performance is promising for large logs with $3\cdot10^7$ events (less than one second) and for very large logs with $3\cdot10^8$ events (less than 20 seconds). 
As expected, the until operator impacts also the performance of more complex formulae (last column in Table~\ref{tab:resBiggestLog}). However, by combining (in conjunction) formulae $\G(\fluent{a} \rightarrow \X\fluent{b}) \wedge \G(\fluent{b} \rightarrow \X\fluent{c})$ and $\fluent{a}\U\fluent{b}$, the execution time increases by only 1.5 times. 
This suggests that complex \FLTLF formulae not involving the until operator are checked very efficiently. In this case, the execution time is around one second for the log with $3\cdot10^7$ events, and around 9 seconds for the log with $3\cdot10^8$. 
%

\section{Conclusions}
We have brought forward a foundational framework, paired with an effective and efficient implementation, for declarative conformance checking of fuzzy event logs. In these logs, events refer in principle to all the possible activities, indicating the intensity of their execution, that is, measuring how much the agents participating to the process engage in their execution. This interpretation is compatible with the relevant setting in which event logs are not explicitly stored inside an information system, but are instead inferred from event extraction and recognition (typically, machine learning-based) pipelines applied to raw multimedia data (such as video). 
Our approach substantiates conformance checking in this setting by pairing \LTLF with fuzzy boolean operators to express state formulae on events, and by computing the fuzzy truth value of a given temporal formula, evaluated on fuzzy traces. This logic corresponds to FLTL \cite{FrigeriPS14} but over finite traces, and lends itself to an efficient, parallel implementation based on PyTorch, to simultaneously check fuzzy conformance of multiple traces. Once interpreted on  crisp logs, the logic behaves as standard \LTLF, so as a by-product we have a very efficient implementation of \LTLF standard (crisp) conformance as well.

Our approach opens up several research directions. We mention four. First, a natural extension is to consider temporal operators that are fuzzy, thus supporting fuzziness not only at the event level, but also across time. To this end, fuzzy-time linear extensions of \FLTLF can be explored, using \cite{FrigeriPS14} as a basis (once suitably re-defined over finite traces). Second, the value obtained as the result of our conformance checking task  can in principle be used in the loss function of a neural network for activity recognition, in the spirit of~\cite{InnesR20,SerafiniGBDSB21}. This woudl yield a neuro-symbolic system that, during training, takes into account the training data as well as background temporal knowledge on the process. This types of systems have been proved to overcome purely data-driven approaches in many scenarios~\cite{SerafiniGBDSB21}. Third, we envision the possibility of combining different forms of uncertainty (such as fuzziness and probabilities), in turn providing a spectrum of uncertainty-aware conformance checking techniques that can be interconnected with machine learning event recognition pipelines, under different assumptions on how events are extracted. Lastly, we aim to integrate the checker in the well-known Declare4Py~\cite{DonadelloRMS22} Python library for process mining.

\begin{credits}
\subsubsection{\ackname} 
Marco Montali has been partially supported by the PRIN MIUR project PINPOINT Prot.~2020FNEB27 and the nextgenerationeu FAIR ``Integrative AI'' project MAIPM. Ivan Donadello and Fabrizio Maria Maggi have been partially supported by the Free University of Bozen-Bolzano with the PRISMA project.
\end{credits}

\bibliographystyle{splncs04}
\bibliography{bibliography}

\end{document}